\documentclass[runningheads]{llncs}
\usepackage{geometry}
\usepackage{graphicx,float}
\usepackage{cite}
\usepackage{amssymb,amsfonts}
\usepackage{algorithmic}
\usepackage{textcomp}
\usepackage{xcolor}
\usepackage{subfig}
\usepackage{hyperref}
\usepackage{multirow}
\usepackage{changepage}

\begin{document}

\title{Supervised classification of dermatological diseases via Deep learning}

\author{Sourav Mishra\inst{1,3}\and Toshihiko Yamasaki\inst{1}
\and Hideaki Imaizumi\inst{2}}
\authorrunning{S. Mishra et al.}

\institute{The University of Tokyo, 7-3-1 Bunkyo-ku, Tokyo\\ 
\email{\{sourav,yamasaki\}@ay-lab.org}
\and
exMedio Inc., Chiyoda-ku, Tokyo\\
\email{imaq@exmed.io}\\
\and
Microsoft Research Asia, Beijing\\}
\maketitle

\begin{abstract}
This paper introduces a deep learning based classifier for nine prevalent 
dermatological conditions. It is aimed at people without easy access to skin specialists. 
We report approximately 80\% accuracy, in a situation where primary care doctors 
have attained 57\% success rate. Our design rationale is 
centered on deploying it on hand-held devices in near future.
With a shortage of dermatological expertise being observed in several countries and 
disease prevalence in every population sample, machine learning 
solutions can augment medical services. 
Our current attempt establishes that deep learning based techniques are viable avenues 
for preliminary information.

\end{abstract}

\section{Introduction}
Access to quality health services is an established need today. Timely treatment can 
alleviate many medical issues. According to estimates by National Institutes of Health (NIH) 
in US, one out of five Americans could develop a serious dermatological anomaly such as 
skin cancer in their lifetimes. If a diagnosis is made early, the survival rate is close to 98\% 
\cite{stern2010prevalence}. Skin diseases such as contact dermatitis and ringworm, although 
not life threatening, are communicable and spread virulently\cite{agbai2014skin,dawes2016racial}.
At a time when demand for dermatological consultation has been rising, there has been a 
consistent under-supply of dermatologists in many countries. The number of practitioners 
in US has plateaued at 3.6 doctors per 100,000 people\cite{kimball2008us}. Japan is actively 
advocating use of telemedicine in areas which are not well serviced
\cite{imaizumi2017hippocra, dekio2010usefulness, lanzini2012impact}. Because of shortage 
of specialists, immediate medical attention is often provided by general 
practitioners. Lowel et al. have argued that a general practitioner's diagnosis 
is concurrent with a dermatologist's opinion only 57\% of the time \cite{lowell2001dermatology}. 
It is difficult to diagnose a wide spectrum of diseases by classic rule based approaches.
In such circumstances, machine learning aided techniques can be feasible means to apprise 
subjects of possible skin problems.

We attempt to provide such a solution which can indicate a subject if it is required to 
seek consultation urgently. It can also help doctors expedite 
consultancies based on the indicated detection. The mode of information exchange envisioned 
is via smartphone app(s) which can securely relay essential patient information. Our submission 
highlights the development of deep learning (DL) based method embedded at the core of 
this process.

\begin{figure}[t]
	\begin{minipage}[b]{1.0\linewidth}
		\centering
		\centerline{\includegraphics[width=9cm, trim={5mm 25mm 15mm 0mm}, clip]{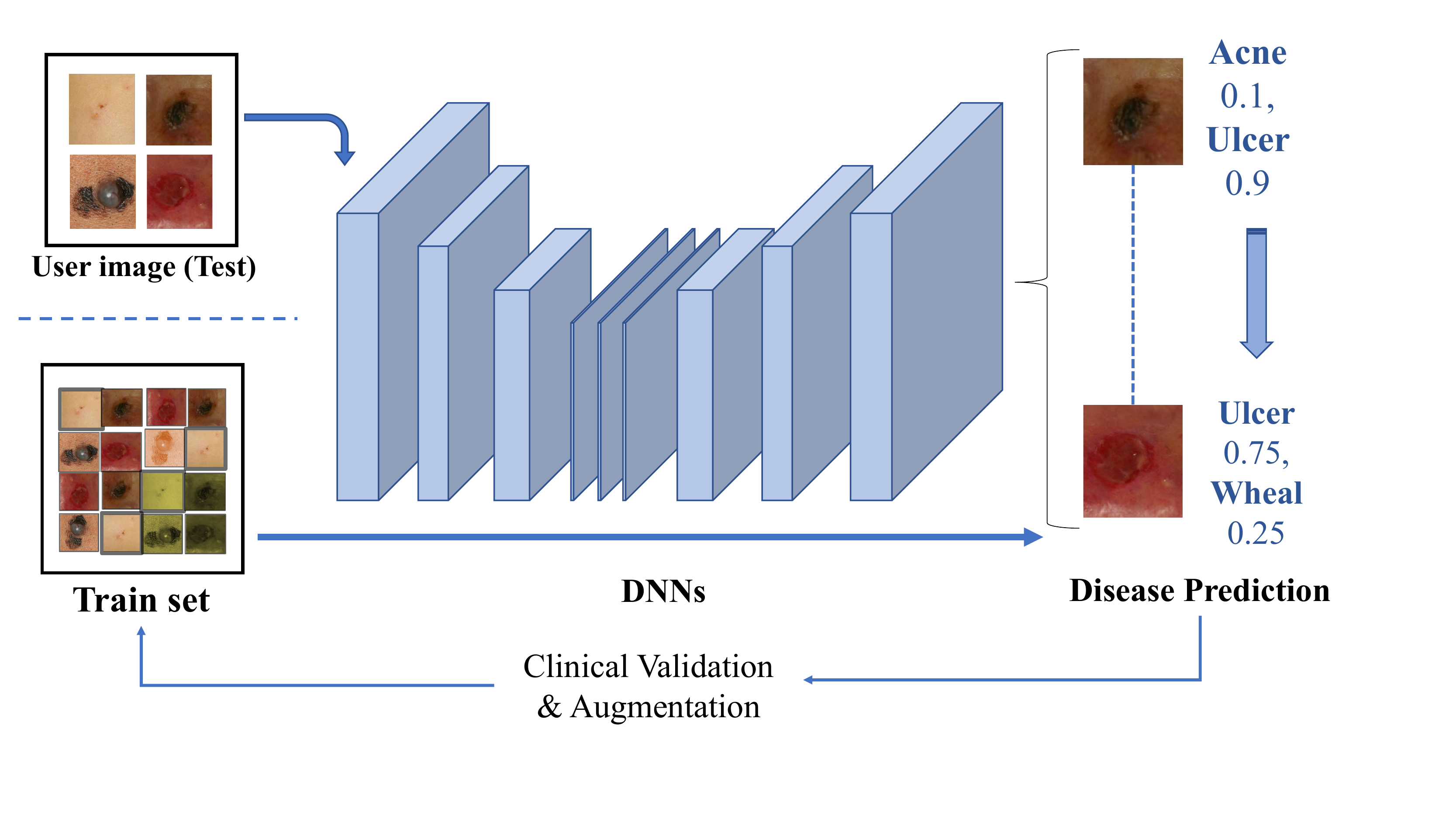}}
		\caption{Schematic of the deep learning based dermatological disease classifier. User 
		supplied test images are classified into one of nine diseases for which the network has 
		been trained. A successful prediction can forewarn the user if there is any urgency in 
		seeking medical attention. The data is vetted by medical practitioners and added back to 
		the training corpus.}
		\label{fig:schema}
	\end{minipage}
\end{figure}

Previously Esteva et al. used deep learning in detecting skin cancer
\cite{esteva2017dermatologist}. Although their research was able to detect \textit{Melanoma} 
with a dermatologist-level accuracy around 74\%, it was limited to skin cancers, and distinguishing 
malignant from benign ones. Similar projects have been conducted by Shrivastava et al. 
in detecting \textit{Psoriasis} \cite{shrivastava2015reliable}. In an attempt to detect 
multiple disease, Park et al. have introduced crowd-sourcing for common skin ailments
\cite{park2017crowdsourcing}. Concurrently detecting multiple common skin diseases is 
unavailable along with lack of labeled data. Most of experiments focus on the accuracy, 
but not on training time or update schemes. We focus on detecting nine common 
skin diseases by training on curated data. We also explore the question of accuracy vis-a-vis 
time to make practical delivery schemes. With human-level accuracy in few classes, we 
hope such methods can gain traction towards affordability in health.

This paper is structured as follows: We discuss data preparation in Section~2.
Our methodology is covered in Section~3. In Section~4, we elaborate on results and conclude 
with a brief discussion on shortcomings \& future directions. The contribution of this 
paper is as follows:
\begin{itemize}
	\item We have curated an image database of nine common dermatological diseases. 
	It comprises of about 4700 images per label. The process isolated only one disease per 
	image. A significant amount of this database is being released to contribute to the
	machine learning community involved in healthcare.
	
	\item We have evaluated classification strategies on popular DNN's fine tuned to our requirements 
	and subsequently attempted to understand the results. 
	
	\item We have compared network schemes and associated training times, which are indirectly 
	related to cost of operation and drawn insights for future work.
\end{itemize}

\section{Data Preparation}
Since disease manifestation in Asian skin types could present differently from other races, 
we performed a systematic data collection via a smartphone application from volunteers. After 
anonymizing, 150,000 clinical images were labeled by trained medical professionals. Nine 
common conditions were chosen from this repository based on prevalence and relevancy.
These diseases were: (i)~\textit{Acne}, (ii)~\textit{Alopecia}, (iii)~\textit{Crust}, 
(iv)~\textit{Erythema}, (v)~\textit{Leukoderma}, (vi)~\textit{Pigmented Maculae}, 
(vii)~\textit{Pustule}, (viii)~\textit{Ulcers} and (ix)~\textit{Wheal}. To avoid any skew in the 
process and balance the dataset, we performed augmentation on the dataset by standard techniques
\cite{russakovsky2015imagenet}. Making judgments from ablation studies we chose 4600 images per 
label approximately. The division of data between training and validation was done in a ratio 
of 90:10. A small corpus of test images, separate from the training and validation set, was kept at 
the outset to assess the classification as a blind experiment. Table~\ref{tab1} illustrates 
information about the various labels and their sizes. 

\begin{table}[t]
	\centering
	\caption{Distribution of image samples across chosen dermatological labels}\label{tab1}
	\begin{tabular}{|c|c|c|c|c}
		\hline
		\textbf{Disease} & \textbf{Training} & \textbf{Validation} & \textbf{Test} \\
		\hline
		\textit{Acne}               &  4215  &  446     & 74 \\
		\textit{Alopecia}	  	    &  4119	 &  441     & 65 \\
		\textit{Crust}		    	&  4147  &  402     & 53 \\
		\textit{Erythema}     	    &  4299  &  406     & 59 \\
		\textit{Leukoderma}         &  4300  &  403     & 58 \\
		\textit{P. Maculae} 	    &  4300  &  310     & 58 \\
		\textit{Pustule}		    &  4046  &  386     & 55 \\
		\textit{Ulcer}				&  4514  &  395     & 58 \\
		\textit{Wheal}		   		&  4120  &  385     & 50 \\
		\hline
	\end{tabular}
\end{table}

\section{METHODOLOGY}
\subsection{Statistical Basis}
Our goal was to get the probabilistic predictions of the diseases as close as possible 
to ground truth. We chose to minimize cross-entropy loss as the basis of a good classification. 
Further information on them can be found in standard literature on statistical methods.

In addition to accuracy we paid attention to training time. Our long term objective requires 
us to frequently retrain models with new data. Training networks from scratch was found to be 
inefficient with best validation accuracy of less than 45\%. We explored popular pre-trained 
DNNs such as ResNet18, ResNet50, ResNet152 and DenseNet161, initialized on ImageNet, as starting 
points for transfer learning\cite{he2016deep,iandola2014densenet,deng2009imagenet}. 
Two strategies were evaluated. The first consisted of tuning the last fully-connected layer 
of these DNNs. The second approach was more rigorous by fine-tuning the entire network. 

The classifier was built on PyTorch (v0.4) framework with Skorch library for scikit-learn 
modules. We chose a batch size of 16 and Stochastic gradient descent (SGD) with a learning rate 
of 0.001 along with appropriate decay for optimizer. The task was run on a system running NVIDIA 
Titan XP and CUDA~v8. Five-fold cross-validation was adopted to deter over-fitting in addition. 
Best weights were recorded as soon as validation loss stabilized by Early-stopping.

\subsection{Training the DNN's}
To test the first approach, we froze the network except for the final fully connected (FC) layer.
Gradients were not computed in the backward direction, so as to not disturb the preceding layers. 
The results obtained were unsatisfactory in comparison to a full training, with maximum validation 
accuracy of 68\% on any of the aforementioned model. We adopted training the full network for 
classification, although it was comparatively much slower. We highlight the results of this step in 
Table~\ref{tab2}. 

\subsection{Test of classification}
530 images, uniformly distributed across the nine labels, were left out of the training \& 
validation corpus. Serving as unlabeled data, they were used to evaluate the quality of 
classification from our fine-tuned DNNs. To accomplish this step, a forward pass of the images 
on networks initialized with the corresponding best parameters was performed. The output score 
indicated the degree of match with each label. These outcomes were matched against the actual 
disease information tabulated by medical specialists. The average time to predict the class for a 
sample was approximately 0.4 seconds without needing a GPU. Test results have been elaborated in 
Table~\ref{tab3}. 

\begin{table}[t]
	\centering
	\caption{Peak Training, validation accuracy \& training times}\label{tab2}
	\begin{tabular}{|c|c|c|c|c|}
		\hline
		\textbf{Network} & \textbf{Validation} & \textbf{Time (min)}\\
		\hline
		ResNet18               &   77.39\%     &   140.50\\
		ResNet50   	  	       &   78.19\%     &   374.11\\
		ResNet152 		       &   84.38\%     &   840.70\\
		DenseNet161            &   82.19\%     &   837.75\\
		\hline
	\end{tabular}
\end{table}

Noting that ResNet152 performs the best among candidate models, we have illustrated the 
class-wise prediction accuracy by a confusion matrix (Table~\ref{tab4}). 

\begin{table}[t]
	\centering
	\caption{Test accuracy on fine tuned ResNet152}\label{tab3}
	\begin{tabular}{|c|c|}
		\hline
		\textbf{Tuned Network} &  \textbf{Averaged Top-1 Accuracy} \\
		\hline
		ResNet18        &   77.13\%     \\
		ResNet50        &   78.81\%     \\
		ResNet152       &   82.30\%     \\
		DenseNet161		&   79.68\%     \\
		\hline
	\end{tabular}
\end{table}

\begin{table}[t]
\centering
\caption{Classifier performance of a network based on ResNet152.}
\def\arraystretch{0.8}
\begin{tabular}{|c c c c c c c c c c|}
	\hline
	\multicolumn{10}{c}{\textit{Predicted (Rounded}}	\\
		\textit{Actual} & \footnotesize\textbf{Acne} & \small\textbf{Alopecia} & \small\textbf{Crust} & 
		\small\textbf{Erythema} & \small\textbf{Leukoderma} & \small\textbf{P. Macula} & 
		\small\textbf{Pustule} & \small\textbf{Ulcer} & \small\textbf{Wheal}\\
		\hline
		\textbf{Acne}           &  83.8\% & 0\% & 0\% & 10.8\% & 0\% & 2.7\% &0\% & 2.7\% & 0\%  \\
		\textbf{Alopecia}       & 0\% & 90.8\% & 6.2\% & 3\% & 0\% & 0\% &0\% & 0\% & 0\% 		 \\
		\textbf{Crust} 			&  0\% & 0\% & 60.4\% & 3.8\% & 0\% & 30.2\% &0\% & 5.7\% & 0\%  \\
		\textbf{Erythema}       & 0\% & 0\% & 7\% & 80\% & 0\% & 13\% &0\% & 0\% & 0\%  	 \\
		\textbf{Leukoderma}     & 0\% & 0\% & 0\% & 0\% & 93.0\% & 3.5\% &0\% & 0\% & 3.5\%  	 \\
		\textbf{P. Macula}      & 3.6\% & 0\% & 0\% & 0\% & 0\% & 96.4\% &0\% & 0\% & 0\%  		 \\
		\textbf{Pustule} 	    &  14.5\% & 11.0\% & 0\% & 9.0\% & 0\% & 0\% & 65.5\% & 0\% & 0\% \\
		\textbf{Ulcer}          &  0\% & 0\% & 6.9\% & 0\% & 0\% & 0\% &0\% & 93.1\% & 0\%          \\
		\textbf{Wheal}    	    &  0\% & 0\% & 0\% & 28.0\% & 0\% & 0\% &0\% & 0\% & 72.0\%  		  \\
		\hline
	\end{tabular}
	\label{tab4}
\end{table}

\begin{figure}[H]
    \centering
	\subfloat[\textit{Alopecia}]{{\includegraphics[height=0.20\linewidth]{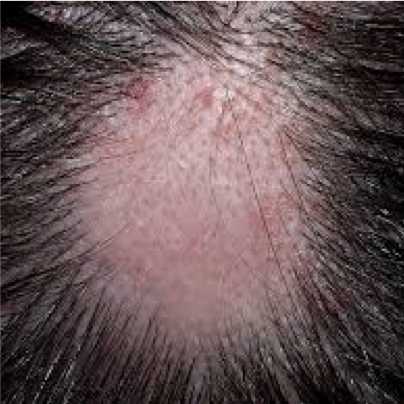} }}%
	\qquad
	\subfloat[\textit{P. Macula}]{{\includegraphics[height=0.20\linewidth]{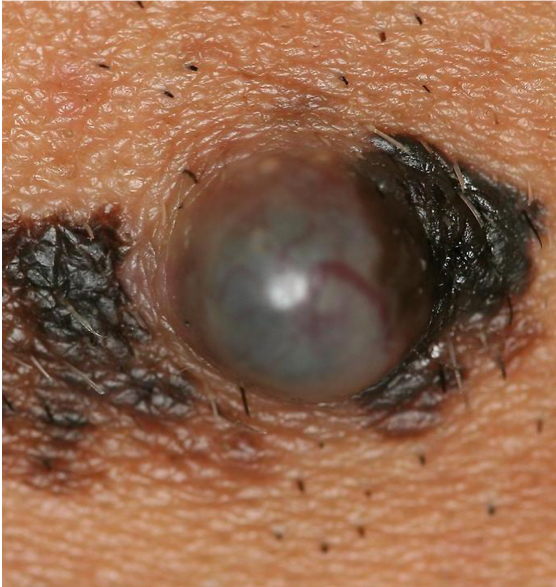} }}%
	\caption{\textit{P. Macula \& Alopecia} exhibit very distinct pattern, structure and contrast}%
\label{fig2}
\end{figure}

\begin{figure}[t]
	\centering
	\subfloat[\textit{Wheal}]{{\includegraphics[height=0.20\linewidth]{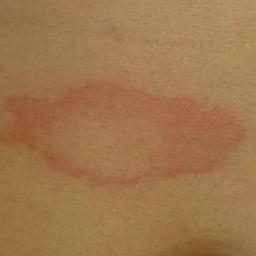} }}%
	\qquad
	\subfloat[\textit{Crust}]{{\includegraphics[height=0.20\linewidth]{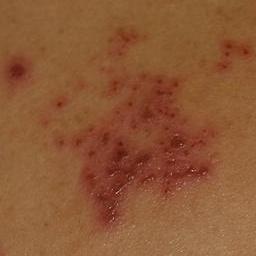} }}%
	\caption{\textit{Wheal \& Crust} often lack high contrast and structure. 
		Such examples present possible ambiguity to the classification process}%
    \label{fig3}
\end{figure}

\section{DISCUSSION}

Despite some detection skew seen in Table~\ref{tab4}, the model performed reasonably well. 
Five classes had test accuracy over 80\%. Further, accuracy below 70\% was observed only in two labels.
Diseases such as \textit{Pigmented Macula, Ulcer} and \textit{Alopecia} are visually distinct in 
terms of contrast and structure. Hence, we hypothesize that extracted features are easy to 
distinguish in such cases. Labels such as \textit{Wheal} or \textit{Crust}, can present difficulty 
because of low amount of texture information in the images. This is consistent with our expectations. 
We illustrate our observation with samples in Fig.~\ref{fig2} and \ref{fig3}. 

From our results, it is abundantly clear that common skin ailments are easy to detect. 
However, there are some caveats we would like to present. We concede that we assumed the existence 
of one of the disease types at the outset. We have not factored in separating normal skin from 
diseased conditions, which is a challenge by itself. Our task was easier than real world scenario, 
where several skin color and types could be involved. Our current results are limited to nine 
commonly seen conditions without any score of the severity. Also, observing the existence of two 
or more disease labels in a single sample is not uncommon. We hope to incorporate solutions to some 
of these situations in future works. In the absence of any network pre-trained on medical skin 
images strictly, our current insights advocate modestly large ResNet architectures when requirements 
deem rapid training and updating necessary. 

\section{CONCLUSION}

This paper elucidates that several common skin problems can be successfully detected with 
deep learning techniques. In absence of dermatologists, this method can predict nine disease 
types, with accuracy surpassing that of general practitioners in many cases. We have also 
highlighted our choice of adopting a particular architecture for further development. 
Although there are some shortcomings, owing to the quantity and complexity of medical images, 
we anticipate overcoming some of these bottlenecks in future experiments. For reproducibility, 
dermatological image data, test codes, along with fine-tuned models are available at 
URL:\url{http://bit.ly/2K76nwx}.

\bibliographystyle{splncs}
\bibliography{ref}
\end{document}